\documentclass{article}
\usepackage{float}

\usepackage{graphicx}
\usepackage{graphicx,wrapfig}
 \usepackage{amsmath}
\usepackage[preprint]{neurips_2024}

\usepackage{pgfplots}
\pgfplotsset{compat=newest}

\usepackage{lipsum} 
\usepackage{algorithm}
\usepackage{algorithmic}
\usepackage[utf8]{inputenc} 
\usepackage[T1]{fontenc}    
\usepackage{hyperref}       
\usepackage{url}            
\usepackage{booktabs}       
\usepackage{amsfonts}       
\usepackage{nicefrac}       
\usepackage{microtype}      
\usepackage{xcolor}         

\usepackage[utf8]{inputenc} 
\usepackage[T1]{fontenc}    
\usepackage{hyperref}       
\usepackage{url}            
\usepackage{booktabs}       
\usepackage{amsfonts}       
\usepackage{nicefrac}       
\usepackage{microtype}      
\usepackage{xcolor}         
\usepackage{amsmath} 
\usepackage{graphicx}
\usepackage{booktabs}
\usepackage{tabularx}
\usepackage{multirow}
\usepackage{xspace} 
\usepackage{multicol}
\usepackage{textcomp}
\usepackage{colortbl}  
\usepackage{xcolor}  
\usepackage{array}
\usepackage{subcaption}

\title{Enhancing Anomaly Detection Generalization through Knowledge Exposure: The Dual Effects of Augmentation}

\author{
  Mohammad Akhavan Anvari \\
  Institute for Research In \\Fundamental Sciences\\
  \\
  \texttt{} \\
   \And
  Rojina Kashefi
 \\
  Institute for Research In \\Fundamental Sciences\\
  \And
  Vahid Reza Khazaie
 \\
  Vector Institute \\
 \AND  Mohammad Khalooei \\
Amirkabir University of Technology
  \\
 \And
  Mohammad Sabokrou \\
Okinawa Institute of Science and Technology \\
}

\begin{document}

\maketitle

\begin{abstract}

Anomaly detection involves identifying instances within a dataset that deviate from the norm and occur infrequently. Current benchmarks tend to favor methods biased towards low diversity in normal data, which does not align with real-world scenarios. Despite advancements in these benchmarks, contemporary anomaly detection methods often struggle with out-of-distribution generalization, particularly in classifying samples with subtle transformations during testing. These methods typically assume that normal samples during test time have distributions very similar to those in the training set, while anomalies are distributed much further away. However, real-world test samples often exhibit various levels of distribution shift while maintaining semantic consistency. Therefore, effectively generalizing to samples that have undergone semantic-preserving transformations, while accurately detecting normal samples whose semantic meaning has changed after transformation as anomalies, is crucial for the trustworthiness and reliability of a model.
For example, although it is clear that rotation shifts the meaning for a car in the context of anomaly detection but preserves the meaning for a bird, current methods are likely to detect both as abnormal. This complexity underscores the necessity for dynamic learning procedures rooted in the intrinsic concept of outliers. To address this issue, we propose new testing protocols and a novel method called Knowledge Exposure (KE), which integrates external knowledge to comprehend concept dynamics and differentiate transformations that induce semantic shifts. 
This approach enhances generalization by utilizing insights from a pre-trained CLIP model to evaluate the significance of anomalies for each concept. Evaluation on CIFAR-10, CIFAR-100, and SVHN with the new protocols demonstrates superior performance compared to previous methods, validating the effectiveness of our approach.

\end{abstract}

\section{Introduction}

Anomaly detection aims to identify samples that significantly deviate from the training data distribution. This task is crucial for the development of reliable machine learning systems and has various critical applications, such as marker discovery in biomedical data \cite{schlegl2017unsupervised} and video surveillance \cite{luo2017revisit}. Unlike traditional classification approaches, which require a well-sampled outlier class, anomaly detection must function with poorly sampled or nonexistent outlier classes. This necessitates the use of one-class classification methods that model the distribution of inlier data to detect outliers \cite{zimek2012survey, sabokrou2018adversarially, Zaheer_2022_CVPR, jewell2022one}.

\begin{figure}[t]
    \centering
    \includegraphics[bb=0 0 1300 600, width=0.9\columnwidth, height=5cm]{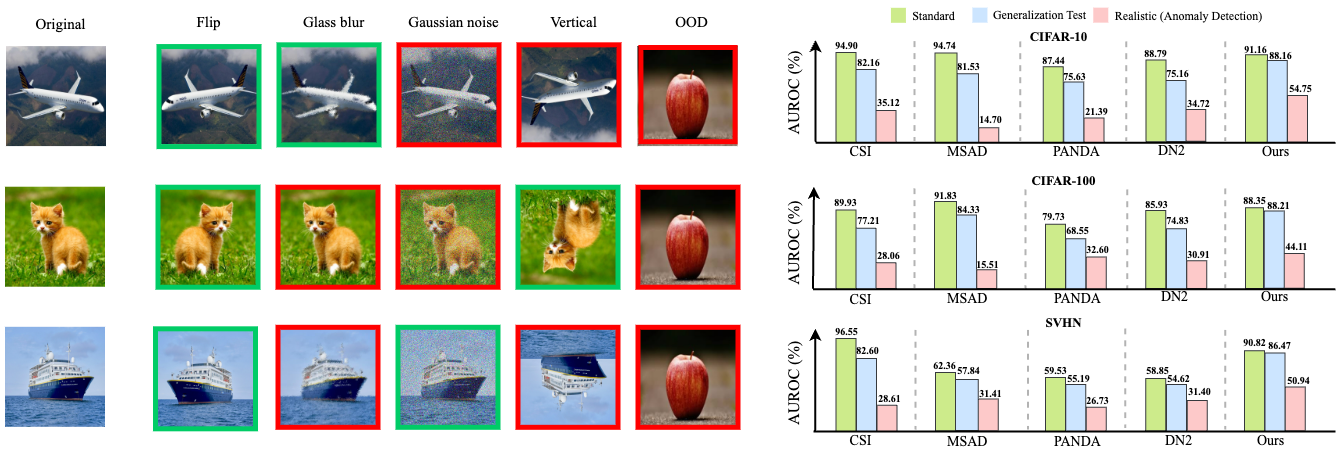}

       \caption{Comparative study on standard and our proposed anomaly detection setups (generalization test and realistic anomaly detection) using CIFAR-10 \cite{krizhevsky2009learning}, CIFAR-100 \cite{krizhevsky2009learning}, and SVHN \cite{netzer2018street} datasets (details of the setups in Section \ref{exp_res}). The left side shows images under various transformations, with red borders indicating negative/anomaly pairs and green borders indicating positive/non-anomaly pairs, selected dynamically for each class through Knowledge Exposure (KE). The right side presents AUROC percentages for different methods, emphasizing that while other methods suffer significant performance drops on more realistic setups due to overfitting and lack of generalizability, our method consistently maintains higher AUROC scores. This demonstrates the robustness and superior effectiveness of our approach in dynamic anomaly detection through KE.}
       \label{fig:f1}
\end{figure}

Current test benchmarks for anomaly detection methods exhibit significant inductive biases that undermine their effectiveness in real-world settings. These benchmarks often assume that normal samples have a distribution very similar to the training set during test time, while anomalies are distributed much further away. This assumption does not hold true in practical scenarios, where real-world test samples often contain various levels of distribution shift which can maintain or change the semantic concept of the sample. This inductive bias leads to flawed testing protocols that encourage methods to have a strong bias towards a low level of diversity in normal data, which is detrimental to their real-world deployability. For instance, state-of-the-art (SOTA) methods like CSI \cite{NEURIPS2020_8965f766} place the decision boundary extremely close to the in-distribution samples, making it easy to detect far out-of-distribution samples but ignoring variations within the inlier sets. As a result, under realistic conditions, these methods fail to develop reliable anomaly detection capabilities. To address this issue, future research needs to shift focus from merely improving performance on existing benchmarks to developing methods that can generalize well to more realistic conditions.

Anomaly detection methods typically use the same approach for all concepts or classes during the training. Usually, methods follow a similar pattern in learning the intrinsic idea of normalcy. For instance, in a dataset with multiple classes, samples are collected uniformly, and even in some methods, training samples are augmented before learning to enhance the learning process. While this approach is logical for balanced image classification, it may not be suitable for anomaly detection, especially with augmentation, as it could alter the normality or abnormality of samples. We aim for our model to handle different types of data augmentation while maintaining its performance and generalization. However, in anomaly detection, some forms of augmentation can change the meaning of the class and cause the samples to be labeled as abnormal. Therefore, the detector must recognize when a particular type of transformation causes the samples to be classified as anomalous while still handling others effectively.

To train an anomaly detector, creating pseudo-abnormal samples from normal ones is a common solution, as often there are no actual labeled abnormal samples during training. Different approaches are used, such as generative models, which can be computationally expensive \cite{mirzaei2022fake, pourreza2021g2d}. Another solution is the generation of pseudo-labels by transforming inlier samples through methods like rotation \cite{NEURIPS2020_8965f766, sohn2020learning}. Some other methods use external datasets to expose the model to abnormal samples and train a detector \cite{hendrycks2018deep}. However, while these methods might work well, none accurately learn the decision boundary required to identify subtle discrepancies around boundaries. Augmentation-based methods that generate abnormal pseudo-samples are better at generating boundary samples, but selecting the transformation to generate pseudo-anomaly samples is challenging. For example, rotation doesn't necessarily change the semantics of an object. For some classes, such as apples, the original apple is considered normal, while the rotated one is considered abnormal, which is nonsensical. This leads to a huge inductive bias, where the performance of rotation detection has a linear correlation with the anomaly detection task, resulting in very low generalization. Moreover, the models are not robust to simple transformations. \textit{This paper addresses the key question of which transformations preserve semantics and should be robust, and which should be detectable as anomalies. It also explores how a model can automatically determine during training whether transformed samples retain their semantics, allowing for the adjustment of positive and negative pairs in conservative learning for anomaly detection.} Answering this question improves the generation of abnormal pseudo-samples, leading to better decision boundaries, enhanced generalization, and more human-like performance.

As aforementioned, certain transformations, such as rotation, can have dual effects: leading to semantic shifts in some objects while not affecting others. Recent research \cite{wang2023rotation} demonstrates a linear correlation between rotation and semantic alterations, although the assumption of anonymity is not universally valid. Additionally, efforts are made to enhance model robustness during testing against convolutional shift variations like Gaussian noise \cite{bai2023feed}. Developing a robust model to handle non-semantic shifting and accurately identifying transformed semantic samples presents challenges, particularly when relying on human supervision. The key aspect is to discern whether an object is sensitive to specific transformations, termed non-transformation-agnostic, or whether it remains invariant despite changes, termed transformation-agnostic. For instance, while cats are commonly observed in various orientations, cars rotated at 90 degrees are less frequently encountered in real-world scenarios. Thus, cats are considered transformation-agnostic, whereas cars are non-transformation-agnostic. The main hurdle arises from the inherent bias in training data, which often lacks the breadth of real-world variations required to recognize the primary functionality amidst different transformations.

To address this issue, we propose the "Knowledge Exposure" method, leveraging the capabilities of the CLIP \cite{radford2021learning} model, which has been trained on a vast dataset. The CLIP's understanding of whether a concept remains invariant across different transformations serves as a valuable asset. If a concept is transformation-agnostic, the CLIP represents both the original and transformed versions similarly, owing to its exposure to a diverse range of transformations during training. Conversely, if a concept is non-transformation-agnostic, the representations differ, as the CLIP encounters fewer instances of the transformed version (see Figure \ref{fig:f1}). This insight guides our approach, which draws inspiration from reconstruction methods \cite{xu2015learning, sabokrou2016video, sakurada2014anomaly, zhai2016deep, zhou2017anomaly, zong2018deep, chong2017abnormal, gong2019memorizing, perera2019ocgan, abati2019latent, alain2014regularized, zimmerer2018context,cong2011sparse, jewell2022one}. In essence, normal samples, which are frequently encountered, exhibit minimal loss during reconstruction. By employing this principle, we aim to identify and prioritize transformation-agnostic concepts, thereby enhancing the model's robustness against non-semantic shifting.

\section{Method}
\textbf{Motivation:} Contrastive learning is a common approach for anomaly detection and is widely used in the literature \cite{NEURIPS2020_8965f766, sohn2020learning}. As we mentioned, these methods mostly consider rotated in-distribution (ID) samples as negative examples and other samples or, for instance, their flipped versions as positive examples. This approach, however, performs well on standard benchmarks but significantly drops when tested for generalization. This is due to two main problems: (1) Considering a fixed augmentation for all ID samples is flawed, as some augmentations preserve the semantics while others do not, which is not suitable in the context of anomaly detection. For example, the model may mistakenly try to maximize the distance between the representation of a cat and a rotated cat. (2) Such models learn a shortcut between being an anomaly and that fixed rotation, leading to a linear correlation between rotation and anomaly detection \cite{wang2023rotation}. To address these issues, we propose dynamically selecting the augmentation that leads to the positive and negative pairs, thereby avoiding the aforementioned problems. 
\textbf{}
\begin{figure}[t]
    \centering
    \includegraphics[bb=0 0 2000 800, width=1\columnwidth, height=6cm]{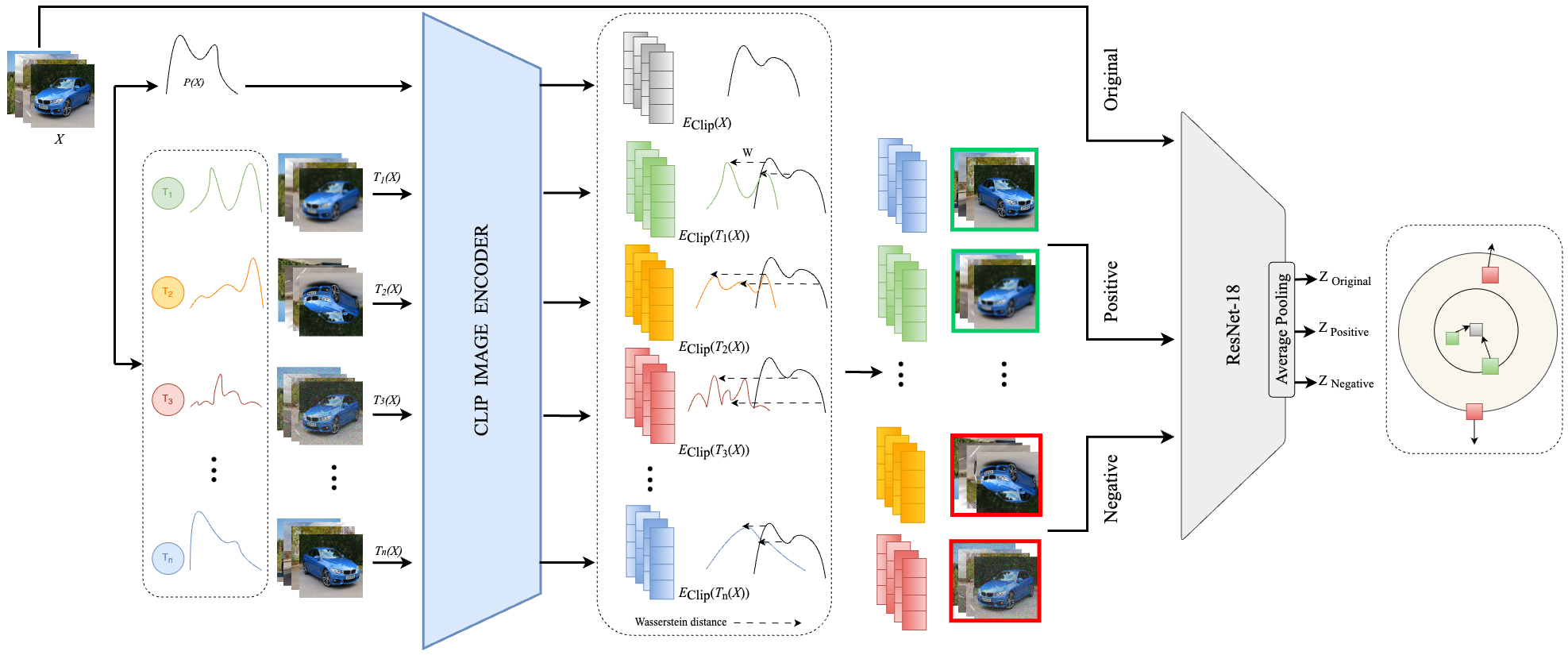}
       \caption{
      The training stage of our anomaly detection framework involves three key steps. First, we apply various geometric and non-geometric transformations to images of each class. Next, we feed both the original and transformed images into the CLIP image encoder, computing the Wasserstein distance between the distribution of the transformed images and the original images from encoder output. Finally, we use the transformations with the smallest distance as positive pairs and those with the greatest distance as negative pairs in contrastive learning. This process ensures effective representation learning for anomaly detection.}
       \label{fig:f2 }
       \end{figure}

\textbf{Problem Definition}
Formally, let \(\mathcal{X}\) be a dataset where the task is anomaly detection, and let \(\mathcal{T}\) be a set of transformations. The goal is to perform supervised contrastive learning, where the positive and negative pairs of each sample are its augmented versions.

For each sample $x \in \mathcal{X}$, we denote its augmented versions under transformations $\mathcal{T}$ as $\mathcal{T}(x) = \{T(x) \mid T \in \mathcal{T}\}$, which either preserve the semantics or change them in the context of anomaly detection concepts.

Positive pairs consist of a sample $x$ and its augmentation $x^+$, where the transformation $T_+ \in \mathcal{T}$ maintains the semantics of $x$. Negative pairs consist of a sample $x$ and its augmentation $x^-$, where the transformation $T_- \in \mathcal{T}$ alters the semantics of $x$ and is contextually unsuitable for anomaly detection. The objective of the supervised contrastive learning task is to maximize the similarity between positive pairs and minimize the similarity between negative pairs, effectively learning representations that distinguish normal samples from anomalies.

The contrastive learning loss for a sample $x \in \mathcal{X}$ is defined as follows:

\[
\mathcal{L}_{\text{contrastive}} = - \log \frac{\exp(\text{sim}(z, z^+)/\tau)}{\sum_{z' \in \mathcal{Z}} \exp(\text{sim}(z, z')/\tau)}
\]

where 
 $z$ is the representation of $x$,
$z^+$ is the representation of the positive pair $x^+$,
 $\mathcal{Z}$ is the set of all representations including positive and negative pairs,
 $\text{sim}(\cdot, \cdot)$ denotes a similarity function (e.g., cosine similarity),
 $\tau$ is a temperature parameter.

This loss encourages the representations of positive pairs to be similar while ensuring that representations of negative pairs are dissimilar. Mathematically, the task is to find transformations $T_- \in \mathcal{T}$ that lead to negative pairs based on the concept, and $T_+ \in \mathcal{T}$ that lead to positive pairs. This process would adapt accordingly for a different dataset $\mathcal{X}'$. Understanding the dynamics of $\mathcal{T}$ based on $\mathcal{X}$, while many of those transformations have never been seen among inlier samples, is very challenging and can even be deemed infeasible. 

\textbf{Knowledge Exposure:} The samples in dataset $\mathcal{X}$ lack the diversity needed to analyze which transformations are typical and which are anomalous. In fact, most samples occur in only one position in the dataset. Understanding which augmentations or transformations preserve the conceptual essence requires a broad perspective and deep insight cultivated by exposure to a wide array of real-world scenarios. Therefore, we are motivated to leverage the CLIP image encoder, which has been trained on 400 million image and text pairs across various contexts. The rationale behind this approach is that CLIP possesses a comprehensive understanding of the real world, enabling it to discern between transformations that maintain the conceptual integrity of an image and those that significantly alter it. By leveraging the extensive training data seen by CLIP, we can identify which augmentations are encountered rarely or frequently during its training, providing valuable insights into their effects on visual concepts.

One of the main paradigms for anomaly detection is reconstruction learning, where it's claimed that learning an encoder-decoder network on dataset \( \mathcal{X} \) will yield low reconstruction error for in-distribution (ID) samples and high error for out-of-distribution (OOD) samples during testing. This concept extends to the latent space, where it's observed that the latent representations of inlier samples are close to each other while those of outliers are far apart \cite{sabokrou2018adversarially}.

Inspired by these observations, we propose leveraging the CLIP image encoder to represent dataset \( \mathcal{X} \) as \( E_{\text{CLIP}}(\mathcal{X}) \). We hypothesize that the distribution of these representations will be close to augmentations that preserve the semantic content and distant from those that alter it significantly. Consequently, we augment the dataset using all transformations \( \{ T_1(\mathcal{X}), \ldots, T_k(\mathcal{X}) \} \) and obtain representations for each augmented dataset with \( E_{\text{CLIP}} \), resulting in \( \{ (E_{\text{CLIP}}(\mathcal{X}), E_{\text{CLIP}}(T_1(\mathcal{X}))), \ldots, (E_{\text{CLIP}}(\mathcal{X}), E_{\text{CLIP}}(T_k(\mathcal{X}))) \} \). 

For a semantic-preserving transformation, we expect each sample from the original dataset to be similar to all transformed images. Measuring the similarity of the entire dataset is necessary, and the Wasserstein distance ($W$) is ideal for this purpose. It quantifies the minimum work required to transform one distribution into another, considering differences in values and positions, and providing a comprehensive understanding of how augmentations affect the dataset. Mathematically, it is formulated as follows:

\[
W(E_{\text{CLIP}}(T_i(\mathcal{X})), E_{\text{CLIP}}(T_j(\mathcal{X}))) = \inf_{\gamma \in \Gamma(E_{\text{CLIP}}(T_i(\mathcal{X})), E_{\text{CLIP}}(T_j(\mathcal{X})))} \mathbb{E}_{(x,y) \sim \gamma} [d(x, y)]
\]

where \(\Gamma(E_{\text{CLIP}}(T_i(\mathcal{X})), E_{\text{CLIP}}(T_j(\mathcal{X})))\) denotes the set of all possible couplings (joint distributions) of \(E_{\text{CLIP}}(T_i(\mathcal{X}))\) and \(E_{\text{CLIP}}(T_j(\mathcal{X}))\), and \(d(x, y)\) is the distance between \(x\) and \(y\). Smaller W-distances indicate positive pairs (where the augmentation does not significantly change the concept from normal to abnormal), while greater distances denote negative pairs. Specifically:

\[
j_{+} = \arg \min_{i} \text{W} (E_{\text{CLIP}}(\mathcal{X}), E_{\text{CLIP}}(T_i(\mathcal{X}))) \quad \{ x, x^{+} \} = \{ x, T_{j_{+}}(x) \}
\]

\[
j_{-} = \arg \max_{i} \text{W} (E_{\text{CLIP}}(\mathcal{X}), E_{\text{CLIP}}(T_i(\mathcal{X}))) \quad \{ x, x^{-} \} = \{ x, T_{j_{-}}(x) \}
\]

This formulation aims to identify which transformations lead to representations that closely resemble the distribution of the original dataset (\( x \)) and which result in representations that deviate significantly (\( x^{-} \)).

In this approach, to achieve better performance, we select \(K\) positive samples, \(\{x^{+}\} = T_{j_+}^{1..K}(X)\), which are the augmentations with the smallest Wasserstein distances from the original data, indicating minimal conceptual change. Similarly, we select \(K\) negative samples, \(\{x^{-}\} = T_{j_-}^{1..K}(X)\), which are the augmentations with the largest Wasserstein distances, indicating significant conceptual changes. Here, \(K\) is a hyper-parameter, and its effect on performance is analyzed in Appendix \ref{sec:app}.

By using these multiple positive and negative samples, the model is trained to maximize the cosine similarity between the anchor \(x\) and its positive samples \(\{x^{+}\}\), while minimizing the cosine similarity between the anchor \(x\) and its negative samples \(\{x^{-}\}\) (see Figure \ref{fig:f2 }). The contrastive loss function \(L_{\text{contrastive}}\) for an anchor \(x\) with multiple positive and negative samples can be formulated as follows:

\begin{equation}
\mathcal{L}_{\text{contrastive}}(x, \{x^{+}\}, \{x^{-}\}) := -\frac{1}{|\{x^{+}\}|} \log \frac{\sum_{x' \in \{x^{+}\}} \exp(\text{sim}(z(x), z(x'))/\tau)}{\sum_{x' \in \{x^{+}\} \cup \{x^{-}\}} \exp(\text{sim}(z(x), z(x'))/\tau)}
\end{equation}

where \(z(x)\) is the representation of \(x\) learned by the model (in this case, we consider ResNet-18).

One question that may arise is why we don't use the CLIP representation directly for anomaly detection. If we were to feed individual samples and their augmentations into CLIP and calculate the distance of their representations from the original dataset, the differences might not be significant enough to detect anomalies reliably. Instead, we compare these differences across the entire dataset to determine whether an augmentation is negative, applying this understanding in the contrastive loss to emphasize it. We will demonstrate this through an experiment in section \ref{exp_res}.

\textbf{Anomaly Scoring:} To calculate anomaly scores using our method, we start by identifying positive (normal) and negative (anomalous) pairs through Knowledge Exposure. We train a ResNet-18 model in a contrastive learning paradigm using these pairs. This involves maximizing the similarity between representations of positive pairs while minimizing the similarity between representations of negative pairs. After training the ResNet-18 model, we extract feature representations for all samples in the dataset. These feature representations are then fed into a one-class Support Vector Machine (SVM), which is specifically trained to distinguish whether a new data point belongs to the same distribution as the training data (normal). The one-class SVM assigns anomaly scores to each sample based on its feature representation. Samples that fall within the learned boundary of normal samples receive lower anomaly scores, indicating they are normal. Conversely, samples that fall outside this boundary receive higher anomaly scores, indicating they are anomalies. The one-class SVM outputs score 1 for normal and -1 for anomaly samples, respectively. The details of the implementation can be found in the appendix.

\section{Experiments and Results}
\label{exp_res}

In this section, we demonstrate how existing anomaly detection methods struggle under our proposed testing protocols, revealing that current anomaly detection benchmarks are inadequate for assessing OOD generalization in real-world scenarios. We illustrate the limitations of these methods in effectively handling semantic-preserving transformations, which are crucial for practical applications. Subsequently, we proposed Knowledge Exposure (KE) method which significantly improves performance and generalization under these rigorous evaluation conditions. Our findings underscore the need for a paradigm shift in the evaluation of anomaly detection models to ensure their reliability and usability in real-world settings.

\begin{figure}[t]
    \centering
    \includegraphics[bb=0 0 700 250, width=0.9\columnwidth, height=5cm]{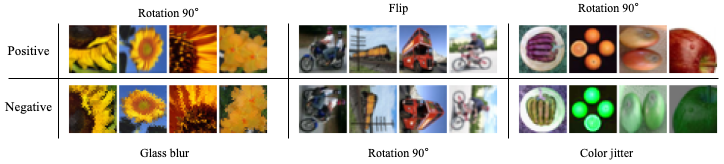}
    \caption{An example of the Knowledge Exposure (KE) mechanism determining positive and negative transformations can be illustrated. For instance, KE identified the flip transformation as positive and the 90-degree rotation as negative for vehicles. Conversely, for fruits, the KE mechanism deemed the 90-degree rotation as a positive transformation and color jitter as a negative one. For sunflowers, a 90-degree rotation is considered positive, while glass blur is considered negative. This selection makes sense because a flip does not change the fundamental appearance of vehicles, whereas a 90-degree rotation does. Similarly, for fruits, color jitter significantly alters their natural perception, while rotation does not. Also, A 90-degree rotation is a fitting positive for sunflowers due to their usual incline to one side, while glass blur which distort their petals, is a negative. Additional images of CIFAR-100 classes can be found in the Appendix ~\ref{sec:app}.}
    \label{fig:f3}
\end{figure}

The Knowledge Exposure (KE) mechanism effectively determined which transformations constituted positive and negative samples for contrastive learning. As depicted in Figure \ref{fig:f3}, for classes such as Bus, Train, and Motorcycle in CIFAR-100, KE identified the flip transformation as positive while recognizing the 90-degree rotation as negative due to the significant alteration of the concept in terms of anomaly detection. Similarly, for fruit images, the KE mechanism deemed the 90-degree rotation as a positive transformation and color jitter as a negative one, aligning well with the inherent semantics of these images. This dynamic selection underscores the capability of KE to discern contextually appropriate augmentations. For more visual results, you can refer to Figure \ref{fig:f5} in the appendix.

For our experiments, we utilize three benchmark datasets: CIFAR-10, CIFAR-100 (focused on coarse-grained setting which contains 20 superclasses), and SVHN (Please see the Appendix for descriptions of these datasets and setups used for generalization evaluation). The results of these experiments will be reported by averaging the Area Under the Receiver Operating Characteristic (AUROC) across all classes. Following the one-class classification approach used in the literature, we evaluate our method in a one-vs-rest setting. The description of the setups we used is as follows:

\textbf{Standard Anomaly Detection (SAD) Setup}: The first setup follows the traditional anomaly detection protocol where one class is designated as normal (inliers), and all other classes are considered anomalies (outliers). This setup, however, suffers from significant flaws and inductive biases. It assumes that normal samples during test time have a distribution very similar to the training set, while anomalies are distributed much further away. In real-world scenarios, test samples often exhibit various levels of distribution shifts while maintaining semantic consistency. This setup makes it easier to detect anomalies that are far from the normal samples but fails to account for the variations within the inlier set. Consequently, methods that simply enhance performance on existing anomaly detection benchmarks without addressing these nuances do not advance the field toward more practical and applicable anomaly detection techniques. The next two setups are designed to evaluate OOD generalization in the context of anomaly detection. 

\textbf{Semantic-Preserving Augmentation (SPA) Setup:} The second setup we aim to create an anomaly detection benchmark that includes sufficient intra-class distribution shift on a wide range of transformations. In this setup, models are expected to be robust to image transformations. We consider images of one class and their transformations as normal samples, while images of other classes and their transformations are treated as anomalies. The reason we used transformations for images of other classes is that current methods create a shortcut between transformation and being classified as an anomaly; by applying transformations to outliers as well, we ensure that the model cannot rely on transformations alone to detect anomalies.

\textbf{Semantic-Shift Aware (SSA) Setup:} The third and most realistic setup we propose addresses the nuanced requirements of real-world anomaly detection. Unlike the second setup, which assumes all transformations preserve the semantic meaning of an image, this approach recognizes that not all transformations do so in the context of anomaly detection. We carefully determine which transformations maintain the conceptual normality of an image and which ones alter its meaning, designating one class and transformations that do not change the semantic meaning as normal. Other classes and transformations that change the semantic meaning of normal images are considered anomalies. This approach more accurately reflects real-world scenarios and better meets the requirements of anomaly detection than existing benchmarks. The transformations used in this setup are flipping, adding Gaussian noise, applying glass blur, JPEG compression, random cropping, 90-degree rotations (rot90), simulating snow, color jittering, 270-degree rotations (rot270), and applying Gaussian blur.


\begin{table}[htbp]
    \centering
    \caption{AUROC (\%) of various anomaly detection methods trained on one-class setting of CIFAR-10 dataset.}
    \vspace{0.3cm}
    \Large 
    \resizebox{1\textwidth}{!}{ 
    \begin{tabular}{p{2.5cm}ccc|ccc|ccc|ccc|ccc}
        \toprule
        \textbf{Classes} & \multicolumn{3}{c}{\textbf{CSI}} &\multicolumn{3}{c}{\textbf{MSAD}} & \multicolumn{3}{c}{\textbf{PANDA}} & \multicolumn{3}{c}{\textbf{DN2}} & \multicolumn{3}{c}{\textbf{Ours}} \\
        \cmidrule(lr){2-4} \cmidrule(lr){5-7} \cmidrule(lr){8-10} \cmidrule(lr){11-13} \cmidrule(lr){14-16}
                          & SAD & SPA & SSA & SAD & SPA & SSA & SAD & SPA & SSA & SAD & SPA & SSA & SAD & SPA & SSA \\
        \midrule
        Plane  & 90.0 & 79.02 & 37.07 & 95.87 & 79.34 & 24.99 & 88.61 & 72.46 & 22.00 & 87.91 & 77.56 & 35.13 & 91.36 & 93.86 & 52.64 \\ \midrule
        
        Car    & 99.13 & 89.53 & 37.40 & 97.90 & 91.42 & 8.23 & 95.69 & 87.58 & 9.45 & 96.15 & 79.42 & 16.67 & 91.56 & 89.22 & 39.40 \\ \midrule
        
        Bird   & 94.46 & 80.27 & 30.89 & 88.90 & 68.76 & 14.80 & 74.61 & 62.22 & 27.29 & 79.43 & 63.38 & 44.40 &  92.46 & 92.08 & 50.43 \\ \midrule
        
        Cat    & 87.42 & 72.99 & 29.62 & 88.35 & 69.43 & 3.35 & 80.0 & 69.15 & 17.27 & 78.22 & 64.88 & 44.90 & 91.32 & 83.21 & 64.71 \\ \midrule
        
        Deer   & 95.48 & 80.39 & 38.97 &  94.34 & 80.67 & 28.03 & 87.66 & 75.45 & 39.16 & 91.13 & 79.60 & 41.50 & 91.03 & 88.58 & 56.59 \\ \midrule
        
        Dog    & 93.23 & 78.03 & 34.24 & 94.86 & 80.68 & 3.54 & 84.61 & 71.28 & 15.02 & 86.03 & 67.74 & 34.05 & 90.22 & 89.88 & 70.05 \\ \midrule
        
        Frog   & 96.27 & 83.01 & 28.38 & 96.60 & 86.49 & 25.10 & 89.58 & 79.94 & 27.23 & 90.43 & 79.45 & 59.54 & 92.38 & 92.61 & 67.21 \\ \midrule
        
        Horse  & 98.94 & 87.79 & 39.37 & 96.51 & 84.60 & 12.59 & 89.17 & 73.01 & 19.60 & 89.74 & 72.38 & 24.86 & 91.59 & 91.36 & 42.38 \\ \midrule
        
        Ship   & 97.90 & 86.25 & 36.48 & 96.68 & 82.39 & 12.70 & 90.33 & 78.63 & 17.75 & 93.89 & 82.66 & 27.92 & 92.21 & 88.70 & 49.73 \\ \midrule
        
        Truck  & 96.26 & 84.42 & 38.84 & 97.40 & 91.56 & 13.72 & 94.15 & 86.66 & 19.15 & 94.99 & 84.57 & 18.32 & 87.51 & 76.62 & 54.39 \\ \midrule
        
        \rowcolor{gray!15} Mean & 94.90 & 82.16 & 35.12 & \textbf{94.74} & 81.53 & 14.70 & 87.44 & 75.63 & 21.39 & 88.79 & 75.16 & 34.72 & 91.16 & \textbf{88.16} & \textbf{54.75} \\
        \bottomrule
    \end{tabular}
    }
    
    \label{tab:ex1}
\end{table}

\begin{table}[ht]
    \centering
    \caption{AUROC (\%) of various anomaly detection methods trained on one-class setting of CIFAR-100 dataset.}
    \vspace{0.2cm}
    \resizebox{1\textwidth}{!}{ 
    \begin{tabular}{p{1.5cm}ccc|ccc|ccc|ccc|ccc}
        \toprule
        \textbf{Classes} & \multicolumn{3}{c}{\textbf{CSI}} & \multicolumn{3}{c}{\textbf{MSAD}} & \multicolumn{3}{c}{\textbf{PANDA}} & \multicolumn{3}{c}{\textbf{DN2}} & \multicolumn{3}{c}{\textbf{Ours}} \\
        \cmidrule(lr){2-4} \cmidrule(lr){5-7} \cmidrule(lr){8-10} \cmidrule(lr){11-13} \cmidrule(lr){14-16}
                         & SAD & SPA & SSA & SAD & SPA & SSA & SAD & SPA & SSA & SAD & SPA & SSA & SAD & SPA & SSA \\
        \midrule
        Class 0  & 87.15 & 74.57 & 30.14 & 92.15 & 96.09 & 12.94 & 76.40 & 66.78 & 32.29 & 81.71 & 73.17 & 31.23 & 88.27 & 88.28 & 50.00 \\ \midrule
        
        Class 1  & 84.92 & 72.53 & 30.90 & 87.41 & 84.49 & 15.18 & 80.48 & 72.18 & 41.01 & 81.99 & 75.55 & 23.95 & 88.49 & 89.12 & 39.55 \\ \midrule
        
        Class 2  & 88.95 & 82.71 & 19.40 & 95.86 & 82.05 & 18.75 & 89.44 & 77.09 & 45.16 & 92.91 & 82.91 & 10.61 & 85.88 & 91.84 & 44.32 \\ \midrule
        
        Class 3  & 84.61 & 74.77 & 25.24 & 91.04 & 82.65 & 5.11 & 68.98 & 58.96 & 18.66 & 82.82 & 64.73 & 42.26 & 88.93 & 85.18 & 35.89 \\ \midrule
        
        Class 4  & 93.01 & 77.35 & 16.92 &  95.49 & 85.36 & 26.39 & 85.80 & 69.70 & 32.05 & 91.64 & 70.10 & 44.93 & 90.14 & 90.00 & 36.13 \\ \midrule
        
        Class 5  & 82.57 & 71.66 & 22.54 & 92.50 & 92.92 & 3.61 & 77.61 & 63.99 & 20.44 & 89.92 & 68.52 & 10.21 & 90.49 & 90.26 & 32.04 \\ \midrule
        
        Class 6  & 92.63 & 78.94 & 24.96 & 92.63 & 82.55 & 2.63 & 78.08 & 70.27 & 17.47 & 88.78 & 73.12 & 59.41 & 88.54 & 89.63 & 46.37 \\ \midrule
        
        Class 7  & 83.90 & 73.48 & 20.07 & 88.78 & 83.14 & 14.71 & 77.16 & 61.74 & 41.54 & 82.84 & 67.21 & 51.00 & 87.18 & 89.81 & 50.00 \\ \midrule
        
        Class 8  & 93.37 & 70.57 & 32.45 & 95.26 & 90.46 & 9.59 & 81.24 & 66.98 & 33.31 & 89.21 & 70.26 & 26.10 & 92.69 & 82.27 & 35.61 \\ \midrule
        
        Class 9  & 95.54 & 87.26 & 35.86 & 91.33 & 88.26 & 23.00 & 80.69 & 84.73 & 32.11 & 87.39 & 87.69 & 17.73 & 78.37 & 92.19 & 54.59 \\ \midrule
        
        Class 10 & 93.59 & 84.12 & 29.22 &  94.32 & 74.97 & 33.87 & 89.01 & 73.25 & 38.87 & 93.86 & 78.74 & 18.67 & 89.11 & 67.50 & 48.56 \\ \midrule
        
        Class 11 & 90.03 & 83.94 & 33.73 &  89.91 & 82.03 & 26.97 & 69.88 & 66.38 & 33.78 & 78.19 & 77.95 & 16.33 & 89.17 & 94.19 & 43.88 \\ \midrule
        
        Class 12 & 91.27 & 78.48 & 28.88 &  90.67 & 79.54 & 5.92 & 78.44 & 68.25 & 36.35 & 80.74 & 81.86 & 34.93 & 83.15 & 84.86 & 33.83 \\ \midrule
        
        Class 13 & 82.32 & 72.88 & 20.14 & 84.26 & 84.30 & 8.79 & 75.48 & 59.15 & 38.30 & 77.00 & 67.94 & 23.93 & 88.84 & 89.12 & 39.44 \\ \midrule
        
        Class 14 & 94.90 & 81.28 & 30.34 & 94.22 & 81.34 & 1.83 & 84.47 & 71.84 & 15.52 & 88.88 & 74.91 & 46.27 & 90.82 & 90.97 & 57.75 \\ \midrule
        
        Class 15 & 85.76 & 69.03 & 28.84 & 86.77 & 83.34 & 11.83 & 69.46 & 61.03 & 41.47 & 80.88 & 77.87 & 39.98 & 89.74 & 92.64 & 39.37 \\ \midrule
        
        Class 16 & 84.13 & 77.12 & 29.70 & 85.28 & 84.45 & 10.90 & 74.00 & 70.75 & 34.73 & 76.39 & 74.30 & 12.34 & 90.89 & 86.19 & 59.24 \\ \midrule
        
        Class 17 & 97.26 & 73.55 & 29.99 & 97.00 & 80.84 & 38.22 & 94.26 & 63.16 & 43.13 & 96.05 & 70.62 & 34.67 & 86.65 & 86.37 & 38.01 \\ \midrule
        
        Class 18 & 96.82 & 77.19 & 37.17 & 97.02 & 84.83 & 20.32 & 84.52 & 71.83 & 25.16 & 88.77 & 79.25 & 40.35 & 95.77 & 97.61 & 44.62 \\ \midrule
        
        Class 19 & 95.99 & 82.80 & 34.79 & 94.79 & 83.08 & 19.69 & 79.22 & 72.95 & 30.69 & 88.72 & 79.91 & 33.38 & 84.04 & 86.18 & 53.03 \\ \midrule
        
        \rowcolor{gray!15} Mean & 89.93 & 77.21 & 28.06 & \textbf{91.83} & 84.33 & 15.51 & 79.73 & 68.55 & 32.60 & 85.93 & 74.83 & 30.91 & 88.35 & \textbf{88.21} & \textbf{44.11} \\
        \bottomrule
    \end{tabular}
    }
    \label{tab:ex2}
\end{table}

\begin{table}[ht]
    \centering
    \caption{AUROC (\%) of various anomaly detection methods trained on one-class setting of SVHN dataset.}
    \vspace{0.2cm}
    \small 
    \resizebox{1\textwidth}{!}{ 
    \begin{tabular}{p{1.5cm}ccc|ccc|ccc|ccc|ccc}
        \toprule
        \textbf{Classes} & \multicolumn{3}{c}{\textbf{CSI}} & \multicolumn{3}{c}{\textbf{MSAD}} & \multicolumn{3}{c}{\textbf{PANDA}} & \multicolumn{3}{c}{\textbf{DN2}} & \multicolumn{3}{c}{\textbf{Ours}} \\
        \cmidrule(lr){2-4} \cmidrule(lr){5-7} \cmidrule(lr){8-10} \cmidrule(lr){11-13} \cmidrule(lr){14-16}
                         & SAD & SPA & SSA & SAD & SPA & SSA & SAD & SPA & SSA & SAD & SPA & SSA & SAD & SPA & SSA  \\
        \midrule
        Digit 0  & 96.85 & 82.49 & 24.03 & 68.16 & 60.80 & 8.10 & 62.16 & 54.95 & 4.25 & 65.14 & 56.25 & 8.10 &  94.29 & 89.97 & 48.16\\ \midrule
        
        Digit 1  & 90.99 & 80.06 & 20.19 & 64.89 & 59.17 & 10.56 & 66.28 & 58.29 & 6.29 & 67.82 & 62.20 & 10.56 & 89.12 & 90.41 & 42.54 \\ \midrule
        
        Digit 2  & 97.43 & 84.26 & 28.66 & 64.48 & 58.80 & 39.05 & 60.35 & 56.55 & 33.02 & 60.31 & 55.73 & 39.05 & 90.32 & 86.01 & 50.48\\ \midrule
        
        Digit 3  & 94.69 & 79.75 & 37.63 & 60.49 & 57.19 & 37.89 & 58.66 & 55.86 & 34.34 & 57.12 & 54.41 & 37.89 & 92.65 & 70.01 & 57.19\\ \midrule
        
        Digit 4  & 98.78 & 83.96 & 33.13 & 65.49 & 58.18 & 50.43 & 63.69 & 57.95 & 48.62 & 65.24 & 59.43 & 50.43 & 88.47 & 91.94 & 54.14\\ \midrule
        
        Digit 5  & 97.00 & 83.41 & 31.70 & 60.23 & 58.05 & 40.16 & 57.81 & 55.39 & 34.94 & 60.93 & 55.04 & 40.10 & 88.58 & 79.12 & 54.40\\ \midrule
        
        Digit 6  & 96.82 & 84.56 & 22.12 & 58.11 & 55.76 & 26.57 & 55.70 & 53.06 & 23.53 & 51.78 & 49.87 & 26.57 & 88.90 & 90.61 & 63.33 \\ \midrule
        
        Digit 7  & 98.70 & 82.13 & 31.21 & 57.42 & 53.41 & 32.45 & 60.54 & 55.75 & 30.31 & 60.96 & 56.90 & 32.45 & 93.78 & 93.81 & 44.81\\ \midrule
        
        Digit 8  & 96.71 & 81.68 & 29.06 & 65.26 & 60.18 & 31.27 & 57.19 & 54.00 & 21.37 & 51.27 & 49.69 & 31.27 & 90.66 & 85.10 & 47.64\\ \midrule
        
        Digit 9  & 97.53 & 83.80 & 28.38 & 59.23 & 56.93 & 37.63 & 52.99 & 50.14 & 30.65 & 47.91 & 46.78 & 37.63 & 91.51 & 87.82 & 46.81\\ \midrule
        
        \rowcolor{gray!15} Mean & \textbf{96.55} & 82.60 & 28.61 & 62.36 & 57.84 & 31.41 & 59.53 & 55.19 & 26.73 & 58.85 & 54.62 & 31.40 & 90.82 & \textbf{86.47} & \textbf{50.94}\\
        \bottomrule
    \end{tabular}
    }
    \label{tab:ex3}
\end{table}

Here, we focus on presenting the results from the Semantic-Preserving Augmentation (SPA) and Semantic-Shift Aware (SSA) setups, as they are the main focus of our evaluation. These setups provide a more stringent and realistic assessment of anomaly detection methods. Tables \ref{tab:ex1}, \ref{tab:ex2} and \ref{tab:ex3} summarize the performance of various anomaly detection methods such as CSI\cite{NEURIPS2020_8965f766}, MSAD\cite{reiss2023mean}, PANDA\cite{reiss2021panda}, DN2\cite{bergman2020deep} and our proposed method, on CIFAR-10, CIFAR-100, and SVHN datasets on all setups. The results are reported in terms of AUROC, averaged across all classes. Our method demonstrates significant improvements over existing methods across all datasets and SPA and SSA setups. These results highlight the effectiveness of our method under challenging and realistic evaluation conditions, confirming the benefits of dynamic augmentation selection and contrastive learning framework in enhancing anomaly detection capabilities. While our method provides better generalization, it achieves comparable performance to existing methods on SAD setup. Those methods with slightly higher performance achieve that through sacrificing generalization and overoptimizing on the flaws existing in current testing protocols.

 \begin{table}[ht]
\caption{Using only the raw features from different pre-trained backbones for anomaly detection on the CIFAR-100 dataset. The backbones that start with C are all CLIP image encoders.}
    \centering
    \renewcommand{\arraystretch}{1.2} 
    \Large 
    \resizebox{1\textwidth}{!}{ 
    \begin{tabular}{l c c c c c c c c c c c c c c c}
        \toprule
        \textbf{Backbone} & \textbf{C\_RN50} & \textbf{C\_RN101} & \textbf{C\_RN50x4} & \textbf{C\_RN50x16} & \textbf{C\_RN50x64} & \textbf{C\_ViT-B/16} & \textbf{C\_ViT-B/32} & \textbf{C\_ViT-L/14} & \textbf{ViT-B/16} & \textbf{ConvNeXt/B} & \textbf{CSWin/B} & \textbf{PatchConvnet/B60} & \textbf{VAN/L} & \textbf{PVT/B4} & \textbf{Ours(ResNet-18)} \\
        \midrule
        \textbf{AUROC} & 82.18 & 83.65 & 82.62 & 80.25 & 78.94 & 76.82 & 75.95 & 81.56 & 59.07 & 62.14 & 86.42 & 63.93 & 66.49 & 70.69 & \textbf{88.35} \\
        \bottomrule
    \end{tabular}
    }
    \label{tab:ex4}
\end{table}

\textbf{Raw Features of CLIP for Anomaly Detection:}
We conducted an ablation experiment to assess the performance of using the raw features of the CLIP image encoder directly for anomaly detection. Specifically, we applied a one-class SVM on these raw features to detect anomalies under the SSA and SPA setups. For better comparison, we also included the raw features of other image encoders. The results indicated that using raw CLIP features did not lead to robust performance, yielding significantly lower AUROCs compared to our proposed method. Table \ref{tab:ex4} demonstrates that raw features from the CLIP encoder are not sufficient for effective anomaly detection under our rigorous testing protocols. Our ResNet-18 model, despite having significantly fewer parameters, outperformed all pre-trained networks with many more parameters. This superior performance can be attributed to the ResNet-18 being trained through Knowledge Exposure and further learning the nuances with contrastive learning, which enabled it to achieve better results than the raw features from all pre-trained networks. As mentioned previously, large-scale encoders are generally robust against most transformations.

\textbf{Alternative Encoders for Knowledge Exposure:}
To evaluate the flexibility of our framework, we experimented with different image encoders instead of the CLIP image encoder. We used DINOv2 \cite{nov2023dinov2, wu2023vision} as an alternative encoder for knowledge exposure. The results, evaluated through human supervision, indicated that DINOv2 did not generalize well across all augmentations and transformations, leading to lower performance compared to CLIP. However, our core idea is not restricted to the CLIP model. The framework remains effective with different encoders, although CLIP yielded the best results due to its extensive training on a diverse dataset. The performance was assessed based on the distance obtained by knowledge exposure from both models.

\begin{figure}[ht]
    \centering
    \includegraphics[bb=0 0 600 350, width=0.9\columnwidth, height=5cm]{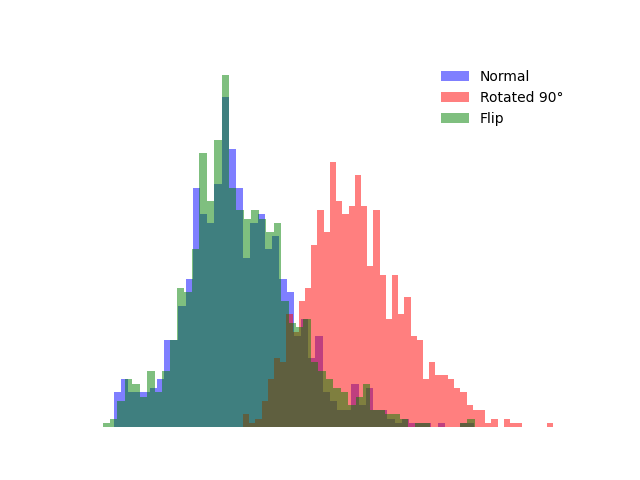}
       \caption{This figure illustrates the distribution of distances under different transformations: the blue histogram represents the original (normal) images, the green histogram represents flipped images, and the red histogram represents 90-degree rotated images of all instances from the "car" class in the CIFAR-10 dataset. The significant overlap between the rotated and normal data indicates that the model has a strong relation to rotation.}
       \label{fig:f4}
\end{figure}

\textbf{The Effect of Transformations on Anomaly Class}:
To specifically address the impact of certain transformations on anomaly detection models, we designed an ablation experiment. We aimed to evaluate how these transformations, when applied to anomaly classes, influence the models' ability to correctly classify anomalies. We observed that the models exhibited a bias, often resulting in false negatives, where anomalies were misclassified as normal due to the presence of similarly transformed normal samples. As illustrated in Figure \ref{fig:f4}, the distribution of distances extracted from the DN2 model for the "Car" class of the CIFAR-10 dataset shows that our assumption of the model forming a strong relation to transformations holds. This figure demonstrates that rotated instances partially overlap with the distribution of normal instances.

\section{Conclusion and Limitations}
In this paper, we address an important shortcoming in the literature on anomaly detection. We demonstrate that each transformation or augmentation has two facets: one that alters the meaning of samples and one that does not, particularly in the context of anomaly detection. Previous methods have overlooked this aspect, resulting in significant limitations in real-world applications and a notable lack of generalization. To understand the dual nature of each augmentation, we leverage the knowledge from a pre-trained model and develop a contrastive learning-based method. This method dynamically selects negative and positive pairs, taking into account both facets of each augmentation. Our results indicate that the proposed method exhibits greater generalization and performs significantly better in realistic scenarios. However, a limitation of our approach is its reliance on a pre-trained network to provide the necessary knowledge for all transformations. This dependency may restrict the method's applicability in situations where such pre-trained models are not available or suitable.

{\small
\setlength{\bibsep}{2pt}
\bibliographystyle{plainnat}
\bibliography{main}
}

\newpage

\appendix
\section{Appendix / supplemental material}
\label{sec:app}
\begin{figure}[h]
    \centering
           \caption{Visualization of dynamically chosen negative and positive pairs for each class through Knowledge Exposure (KE) from randomly selected classes in CIFAR-100.}
\includegraphics[bb=0 0 450 650, width=0.9\columnwidth]{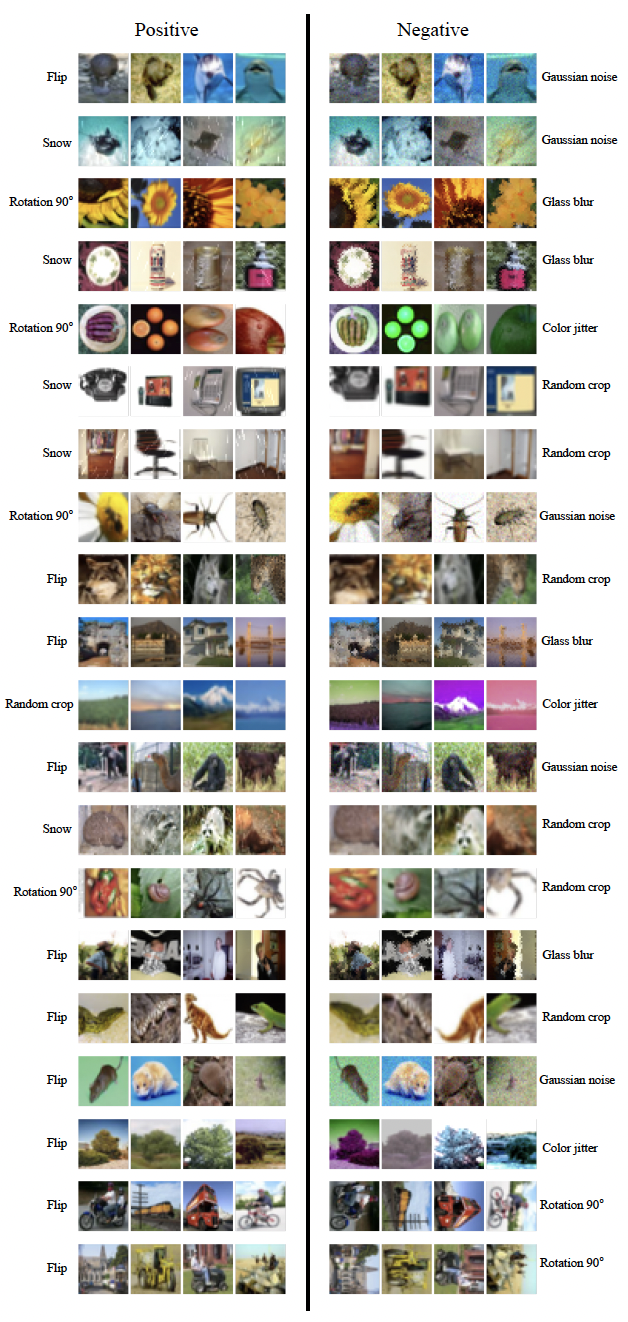}
       \label{fig:f5}
\end{figure}
The effectiveness of Knowledge Exposure (KE) in dynamically selecting appropriate negative and positive pairs is evident. For instance, the identification of color jitter as a negative pair for fruits, sky scenes, or other color-dependent concepts is logical, as these transformations significantly disrupt the color integrity crucial for anomaly detection. Similarly, for objects such as cars, buses, and bicycles, a 90-degree rotation is correctly identified as a negative pair since such rotations would render these objects anomalous in real-world scenarios. Conversely, for objects like apples or oranges, a 90-degree rotation is sensibly selected as a positive pair, as this transformation does not alter their conceptual integrity. Figure \ref{fig:f5} clearly illustrates that KE performs effectively across various cases, reinforcing the results presented in the paper and highlighting the robustness and accuracy of our anomaly detection framework.

\begin{table}[h]
    \centering
    \caption{ The Effect of Different Values of the \textbf{K} Hyperparameter on Model Generalization on CIFAR-10 dataset. The results show that using \textbf{K}=2 enhances the model's generalization by exposing it to two distinct transformations as positive and negative data.}
    \vspace{0.2cm}
    \renewcommand{\arraystretch}{0.1} 
    \small 
    \begin{tabular}{p{1.5cm}|ccc|ccc}
        \toprule
        \textbf{Classes} & \multicolumn{3}{c}{\textbf{K=1}} &\multicolumn{3}{c}{\textbf{K=2}} \\
        \cmidrule(lr){2-4} \cmidrule(lr){5-7} \\
        \midrule
        Class 0 & 75.26 & 81.18  \\ \midrule
        Class 1 & 82.12 & 82.47  \\ \midrule
        Class 2 & 79.11 & 84.33 \\ \midrule
        Class 3 & 79.97 & 81.61 \\ \midrule
        Class 4 & 88.35 & 82.26 \\ \midrule
        Class 5 & 80.78 & 80.68\\ \midrule
        Class 6 & 79.27 & 84.43 \\ \midrule
        Class 7 & 83.43 & 86.89 \\ \midrule
        Class 8 & 77.85 & 82.20\\ \midrule
        Class 9 & 63.03 & 79.41 \\ \midrule
        \rowcolor{gray!15} Mean & 78.91 & 82.54 \\
        \bottomrule
    \end{tabular}
    \label{tab:ex5}
\end{table}

\textbf{Related Works}

One of the most widely used techniques in anomaly detection involves employing self-supervised methods, which generate pseudo-labels from the data itself. Contrastive learning \cite{falcon2020framework}, a self-supervised technique, has shown remarkable success in visual representation learning \cite{he2020momentum, chen2020simple}. This method focuses on learning representations by contrasting positive and negative samples, ensuring that representations of similar (positive) instances are closer together, while representations of dissimilar (negative) instances are further apart. Common loss functions used include triplet loss, NT-Xent loss, and InfoNCE \cite{schroff2015facenet, chen2020simple}. Recent research has indicated that transformations once thought to be harmful in traditional contrastive learning can be beneficial in out-of-distribution (OOD) detection. In the Contrasting Shifted Instances (CSI) method \cite{NEURIPS2020_8965f766}, in addition to contrasting a given sample with other instances, the training scheme also contrasts the sample against distributionally shifted augmentations of itself, which enhances OOD detection \cite{tack2020csi}.Transformations can result in either semantics-preserving or semantics-shifting images, depending on the class, which should be considered when selecting positive and negative pairs. \cite{wang2023rotation} addressed this issue in rotation transformation, but this principle can also apply to several geometric and shifting transformations.

Additionally, several methods use knowledge from pre-trained models to identify normal data patterns. These methods, including DN2\cite{bergman2020deep}, PANDA\cite{reiss2021panda}, and MSAD\cite{reiss2023mean}, first employ the pre-trained models to extract features that represent typical data points. Then, they utilize techniques like k-nearest neighbors (KNN) and Gaussian mixture models (GMM) to assess the distance of new data points from the established set of normal features. This distance is used to calculate an anomaly score.

Furthermore, outlier exposure (OE) is a new technique proposed for anomaly detection tasks, which uses an auxiliary dataset of outliers \cite{hendrycks2018deep}. The main problem with OE is its reliance on a large and diverse outlier dataset during training, which may not be readily available in many practical scenarios. Additionally, the learned representations may not generalize well to unseen outlier distributions, and in the case of irrelevant outliers, performance decreases.

Recently, large pre-trained vision-language models, trained using millions of image-text pairs \cite{radford2021learning}, have demonstrated strong zero-shot recognition ability in various vision tasks, including anomaly detection. These models rely on the ability to transfer knowledge from auxiliary data to identify unseen anomalies. Early approaches, such as CLIP-AD \cite{liznerski2022exposing}, ZOC \cite{esmaeilpour2022zero}, and ACR \cite{li2024zero}, which require tuning on auxiliary data for each target dataset, have been proposed for anomaly classification. Recent approaches focus on both anomaly segmentation and classification. For effective anomaly segmentation, WinCLIP \cite{jeong2023winclip} employs a wide range of hand-designed text prompts and multiple forward passes of image patches. To improve the modeling of local visual semantics, VAND \cite{chen2023zero} introduces learnable linear projection techniques. However, these methods encounter issues because text prompt embeddings lack sufficient generalization, leading to reduced accuracy in identifying anomalies associated with diverse, unseen object semantics. AnomalyCLIP \cite{zhou2023anomalyclip} tackles these challenges by adapting to diverse datasets after being trained on a general dataset. It uses only two trainable, object-independent text prompts for identifying anomalies and segments images with a single forward pass.

However, these methods rely on learning text prompts to capture anomalies, which may not generalize as well as directly learning from image data. Additionally, they use a combination of global and local loss functions, which could be less efficient compared to our adaptive contrastive learning on image features.

\textbf{Implementation Details}

We conducted experiments using the CIFAR-10, CIFAR-100, and SVHN datasets. The CIFAR-10 dataset comprises 60,000 color images of size 32x32, evenly distributed across 10 classes, with each class containing 6,000 images. The CIFAR-100 dataset extends to 100 classes, each with 600 images. Both CIFAR datasets are split into training sets of 50,000 images and testing sets of 10,000 images. The SVHN dataset, derived from Google Street View, consists of over 600,000 32x32 color images of house numbers, used for digit recognition and classification.

In each previously discussed method, we employed standard data and specific configurations unique to each method during their respective training phases. For instance, in the case of CSI, we preserved its fundamental augmentation. We implemented two protocols for every dataset: generalization (SPA) and real-world (SSA). During the generalization training phase, all models requiring training data were provided with a chosen dataset, undergoing its standard pre-processing without modifications or augmentations to the training procedure. In the testing phase, aside from augmenting the test set, no alterations were made to the testing configurations of the methods. For generating augmented datasets we used \cite{khazaie2022out} framework to be as generalized as possible in training datasets we use hyper-parameter severity equal to 1 but in proposed dataset we use 6 for severity to became towards near real world worst cases scenario.

To create the real-world dataset, we used Knowledge Exposure (KE) to generate a list of distances for each original class from various augmentations. These distances were sorted to identify negative augmentations as anomalies. Additionally, in SSA, alongside KE, we employed human supervision to refine the dataset and applied other augmentations and transformations not considered negative in the training process. For example, using KE, Gaussian noise and glass blur were identified as the most negative augmentations for cars. However, in the real world, rotation is considered an anomaly for cars. Therefore, in creating the real-world dataset, we used human supervision to label rotated cars as anomalies.

To create a comprehensive list of anomalies for each class in each dataset, we initially implemented the aforementioned protocols using KE, supplemented by human supervision for accurate and real-world scenario selection. We then loaded the original dataset and iterated through it, applying a randomly selected transformation from our predefined list to each data point. We evaluated whether this transformation was classified as an anomaly based on the previously assembled anomaly transformation list.

For preprocessing the start with augmenting, the original dataset using  and then feed them into the CLIP image encoder to produce the representations. Then use extracted representations in Wasserstein algorithm for scoring each augmentaion.

For training, we utilized ResNet-18 without a classification head as the architecture for our neural network. We trained our model for 50 epochs using the SGD optimizer with a learning rate of 0.01, momentum of 0.9, and a weight decay of 5e-4 for each experiment. We used AUROC as the anomaly detection metric, where the AUROC value ranges from 0 to 1, with values closer to 1 indicating better classifier performance. For loss function we utilize the InfoNCE loss \cite{chen2020simple} with 0.2 temperature value for better convergence rate. Our findings indicate that using 2 for \textbf{K} hyperprameter enhances the generalization which can be interpreted as model being exposed to two distinct augmentations or transformations as positive and negative data which results in constriction of model into more significant features, thereby reducing the likelihood of overfitting to irrelevant features which available in Table \ref{tab:ex5}.

For evaluation we utilized our trained model to extract features from training data for training a One-class SVM with sigmoid kernel. Then for any input data, we use the same model to extract the features and then predict the label, whether input is normal or anomalous.
\end{document}